\documentclass[letterpaper, 10 pt, conference]{ieeeconf}  

\IEEEoverridecommandlockouts                              

\usepackage{amsmath} 
\usepackage{amssymb}  
\usepackage{subcaption} 
\usepackage{graphicx}
\graphicspath{{figure/}}
\usepackage{subfig}
\usepackage[ruled,vlined,linesnumbered]{algorithm2e} 
\DontPrintSemicolon
\SetKwInput{KwRequire}{Require}
\SetKwInput{KwEnsure}{Ensure}
\SetInd{0.5em}{0.8em}            
\usepackage{cite}

\usepackage{xcolor} 
\usepackage{booktabs}	
\usepackage{amssymb}    
\usepackage{array}      
\usepackage{makecell}   
\usepackage{pifont}
\usepackage{tabularx}
\usepackage{multirow}  
\usepackage{float}
\usepackage{balance}

\begin{document}
	
	\title{\LARGE \bf  DiffusionVS: A Generative Framework for Robust Visual Servoing Based on Diffusion Policy}

	\author{Hongkang Cui, Rui He and Haoyao Chen*
	}

	\maketitle
	\thispagestyle{empty}
	\pagestyle{empty}

	\begin{abstract}

		Visual servoing is a fundamental technique in robotic manipulation and navigation. Regression-based visual servoing frequently experiences trajectory jitter as a result of noise-sensitive single-step mappings and the accumulation of errors during distribution shifts. In contrast, Diffusion Policy maintains temporal consistency by predicting action sequences and improves robustness through implicit data augmentation.
		This paper presents a novel diffusion-based servoing method. Based on Diffusion Policy, the proposed approach uses normalized image coordinates of observed tag corners as input and generates camera velocity through conditional denoising. To overcome the generalization limitations of models trained on static datasets, an online training paradigm is adopted, continuously expanding the diversity of training data through interactive experience collection. This strategy substantially enhances both the performance and generalization capability of the model. Comprehensive simulations and real-world experiments demonstrate the effectiveness of the proposed method, achieving success rates of nearly 100\% in simulation and 93\% in physical experiments. Beyond the specific pipeline, we further validate the generality of the diffusion mechanism. Experiments show that existing visual servoing networks consistently achieve improved performance when integrated with our diffusion-based module. These results indicate that the proposed strategy possesses broad applicability and can enhance various visual servoing systems beyond the specific architecture presented here.
		
	\end{abstract}

	\section{INTRODUCTION}
	
	Visual servoing is a fundamental technique in robotics, extensively utilized in manipulation, navigation, and human-robot interaction because of its versatility and effectiveness. Operating as a closed-loop system, it integrates real-time visual feedback into the motion control process. The controller continuously computes the error between the current and desired states, generating motion commands to progressively minimize this error and guide the robot to the target pose.\cite{chaumette2006visual}\cite{hutchinson1996tutorial}
	
	Traditionally, visual servoing methods are classified into two primary approaches: Position-Based Visual Servoing (PBVS) and Image-Based Visual Servoing (IBVS), based on the domain in which the error signal is defined.
	Although PBVS ensures an optimal trajectory in Cartesian space, its effectiveness depends on accurate target depth and precise camera calibration. Uncertainties in system parameters or depth estimation may result in significant steady-state errors or system instability.\cite{wilson1996relative}
	In contrast, IBVS demonstrates strong robustness to camera calibration errors and depth uncertainties.\cite{espiau1992new} However, it faces inherent challenges in mapping two-dimensional image errors to three-dimensional camera motion. This often leads to suboptimal Cartesian trajectories, such as the "camera retreat" problem, where the robot moves backwards unnecessarily to maintain feature visibility. Additionally, IBVS controllers are susceptible to local minima and singularities in the interaction matrix (image Jacobian), which may cause control failures during large displacements.\cite{chaumette1998potential}
	
	Recent advancements in neural network technologies have led to several end-to-end approaches that encode servoing targets into latent vectors and train networks under PBVS-based supervision to predict camera velocity\cite{saxena2017exploring} \cite{bateux2018training}\cite{yu2019siamese}\cite{felton2021siamese}. However, these methods typically employ regression-based architectures that are susceptible to noise-sensitive single-step mappings. Small observation perturbations, especially in proximity to targets, frequently lead to temporally inconsistent actions. This inconsistency results in trajectory jitter and the accumulation of errors, which undermine execution stability and reduce robustness to previously unseen noise distributions.

\begin{figure}[htbp]
		\centering
		\includegraphics[width = 0.5\textwidth]{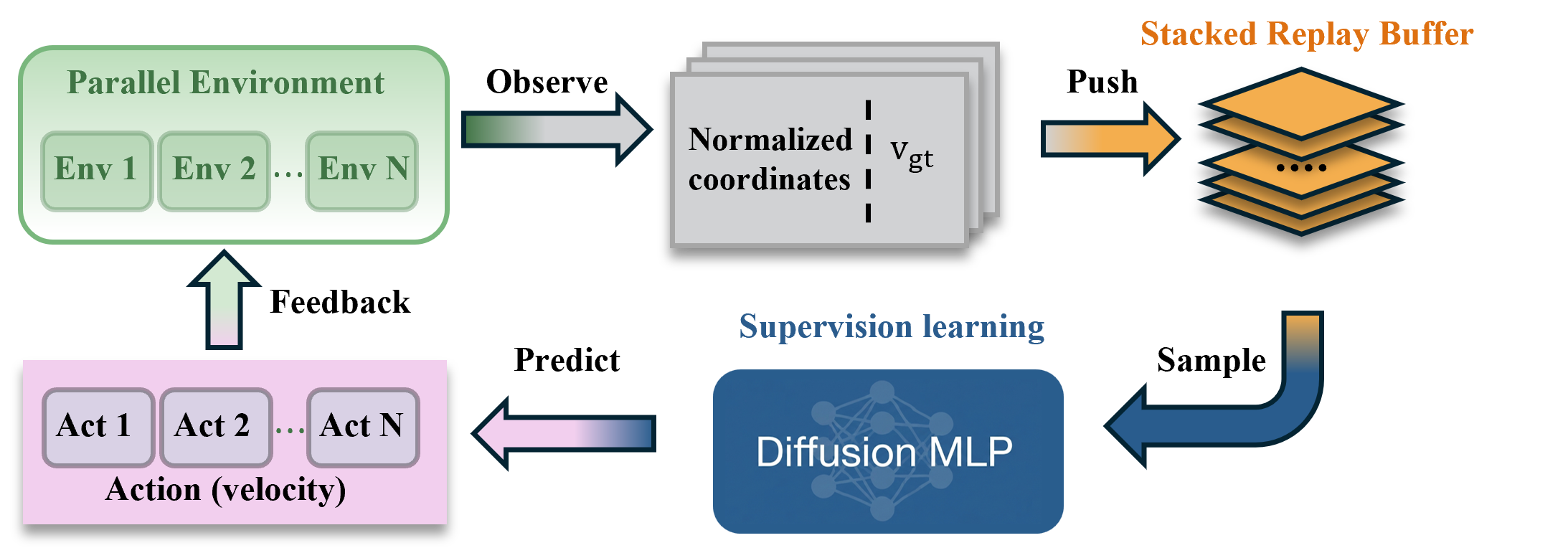}
		\caption{DiffusionVS Framework }
		\label{fig:Overview}
	\end{figure}

	This paper presents a visual servoing method based on Diffusion Policy as illustrated in Fig.\ref{fig:Overview}. Multiple parallel simulation environments are used to continuously collect interactive data during execution, which is stored in a shared replay buffer. Training batches are sampled from this buffer to train the diffusion network. The model learns to denoise velocity sequences and feeds the denoised velocities back into the environment to guide subsequent actions. Ultimately, the trained model receives the current observation as input and transforms a sequence of Gaussian noise into an appropriate servoing velocity through iterative denoising.
	
	The main contributions are as follows:
	
	\begin{itemize}
		\item An end-to-end visual servoing framework, DiffusionVS, is proposed based on the Diffusion Policy. Experimental results demonstrate that this generative approach consistently outperforms comparable regression-based models, achieving a near-perfect success rate of 100\% in simulation (with position and orientation errors of 0.54 cm and 0.53$^\circ$, respectively) and 93\% in real-world experiments (6.17 cm and 3.76$^\circ$).
		\item An online training scheme inspired by reinforcement learning is designed, where the model controls camera motion in simulation based on its predictions. This dynamic strategy is demonstrated to be significantly more effective than static dataset training, expanding data scale and diversity to successfully bridge the sim-to-real gap.
		\item The proposed diffusion module exhibits strong generalizability. We demonstrate that it can serve as a highly effective plug component, boosting the success rates of existing state-of-the-art methods by up to 30\% on specific tasks.
	\end{itemize}

	\section{RELATED WORKS}
	\subsection{Fundamentals of Visual Servoing}
	
	Traditional visual servoing methods encompass Image-Based Visual Servoing (IBVS), Position-Based Visual Servoing (PBVS), and hybrid approaches\cite{he2026cibvs}. IBVS relies on geometric features extracted from images, such as keypoint correspondences, and demonstrates robustness to camera calibration errors and model inaccuracies \cite{chaumette2006visual}\cite{allibert2010predictive}. However, IBVS is limited by a narrow convergence basin and is susceptible to feature loss, unpredictable three-dimensional trajectories, Jacobian singularities, and convergence to local minima \cite{kermorgant2011combining}\cite{jin2021policy}\cite{chaumette2007potential}. In contrast, PBVS uses the camera's three-dimensional pose as the visual feature, providing global asymptotic stability, but it depends on precise camera intrinsic parameters and an accurate three-dimensional model of the observed object for pose estimation. Hybrid methods seek to integrate the advantages of both paradigms by either switching between them or fusing their outputs to achieve more balanced performance \cite{gans2007stable}\cite{malis20022}. Nevertheless, these hybrid strategies do not fully address the fundamental limitations of their underlying paradigms. Ultimately, classical frameworks remain constrained by the narrow convergence basin inherent to IBVS and the strict requirement for accurate three-dimensional information in PBVS.

	\subsection{Learning-Based Visual Servoing}
	
	Recent advancements in neural network technologies have led to the development of various learning-based visual servoing methods, which can be broadly categorized as follows:
	
	The first category utilizes neural networks to enhance the observer component by applying deep learning techniques to improve feature extraction and matching, while maintaining traditional control strategies within the servoing loop \cite{detone2018superpoint}\cite{sarlin2020superglue}\cite{ni2023pats}. While these methods adapt effectively to diverse visual servoing scenarios, they do not fundamentally resolve the inherent limitations of classical control schemes, such as narrow convergence basins, sensitivity to depth estimation errors, and vulnerability to local minima, because the underlying control mechanisms remain unchanged.
	
	The second category focuses on training end-to-end models that directly map input images or feature points to control outputs, such as camera velocity or the interaction matrix \cite{yu2019siamese}\cite{felton2021siame}\cite{10158789}. A prevalent approach encodes the input image or keypoints and employs a lightweight multilayer perceptron (MLP) to predict camera velocity or pose at the subsequent time step \cite{felton2022visual}\cite{urbanikova2023arguing}\cite{Chen2023CNSCE}\cite{Chen2025CNSv2PC}. However, regression-based visual servoing methods often suffer from limited accuracy and stability due to their deterministic, single-step mapping. This makes them sensitive to observation noise, especially near target convergence, where small disturbances can cause errors and trajectory jitter. They are also vulnerable to error accumulation from unseen noise distributions, leading to deviations from expert trajectories. In contrast, generative methods like Diffusion Policy\cite{chi2025diffusion} produce coherent action sequences and maintain temporal consistency, ensuring smoother execution. The noise-injection in diffusion training acts as implicit data augmentation, improving robustness to distribution shifts and observation disturbances.
	
	\subsection{Diffusion Policy}

	Diffusion Policy has recently emerged as a powerful framework for visuomotor learning, demonstrating state-of-the-art performance in complex robotic tasks such as dexterous manipulation and navigation. By effectively modeling multimodal action distributions, it offers superior robustness compared to traditional regression-based methods, building on the remarkable success of diffusion models in generative tasks \cite{maze2023diffusion}\cite{dhariwal2021diffusion}. Given these capabilities, diffusion policies are naturally suited for visual servoing tasks that demand precise control under visual uncertainty. However, their performance heavily relies on the richness of the training data; when trained on static, limited datasets, these models often suffer from insufficient generalization in dynamic or novel environments. Consequently, their application in end-to-end visual servoing frameworks remains largely unexplored.
	
	\section{METHODS}
	\subsection{Problem Definition}
	
	Given a desired image $\mathbf{I_d}$ and the current image $\mathbf{I_c}$, DiffusionVS aims to compute a camera velocity command, expressed in the camera frame, that moves the camera toward the target pose. It does not require exact depth inputs and still generates trajectories of comparable quality to those produced by PBVS.
	
	The observation space is defined on the normalized image plane. In practice, these coordinates are derived from pixel coordinates obtained through visual detection algorithms and are transformed using the camera intrinsic parameters. Let the pixel coordinates of the servoing points (specifically assumed to be AprilTag features in this study) in the image be denoted as 
	
	\begin{equation}
		\mathbf{u}_{k,i} = [u_{k,i}, v_{k,i}]^\top
	\end{equation}
	
	index $k \in \{c, d\}$ indicates the current ($c$) and desired ($d$) states, respectively, and  $i \in \{1, 2, \dots, n\}$ indexes the servoing points. 
	
	Given the known camera intrinsic matrix $\mathbf{K}$, the pixel coordinates are back-projected onto the normalized image plane at $Z = 1$ to obtain the 2D normalized coordinates $\mathbf{x}_{k, i} = [x_{k, i}, y_{k, i}]^\top$. This transformation is defined as follows:
	
	\begin{equation}
		\begin{bmatrix} x_{k,i} \\ y_{k,i} \\ 1 \end{bmatrix} = \mathbf{K}^{-1} \begin{bmatrix} u_{k,i} \\ v_{k,i} \\ 1 \end{bmatrix}
	\end{equation}
	
	Geometrically, these normalized coordinates correspond to the Cartesian coordinates $\mathbf{P}_{k, i} = [X_{k, i}, Y_{k, i}, Z_{k, i}]^\top$ of the servoing points in the camera coordinate system through perspective projection:
	
	\begin{equation}
		\mathbf{x}_{k,i} = \left[ \frac{X_{k,i}}{Z_{k,i}}, \frac{Y_{k,i}}{Z_{k,i}} \right]^\top
	\end{equation}
	
	Consequently, the goal of visual servoing is formalized as establishing a control mapping $f$ from the error space $E$ to the camera velocity space $V$. Defining the error state vector as $\mathbf{e} \in \mathbb{R}^8$, where $\mathbf{e}$ is the difference between the current and desired normalized coordinates, this mapping is expressed as:
	
	\begin{equation}
		f : E \rightarrow V = \{\mathbf{v}_c\}, \quad \mathbf{v}_c \in \mathbb{R}^6
	\end{equation}
	where $E$ represents the space of the error, and $\mathbf{v}_c$ is the 6-DOF camera velocity command generated to drive this error to zero.
	\subsection{Data Preprocessing}
	
	In DiffusionVS, we flatten and concatenate the current and desired feature points to form the observation conditioning vector for the DDPM, given as
	
	\begin{equation}
		\mathbf{cond} = \left[ \mathbf{x}_{c,1}^T, \dots, \mathbf{x}_{c,4}^T, \mathbf{x}_{d,1}^T, \dots, \mathbf{x}_{d,4}^T \right]^T \in \mathbb{R}^{16}
	\end{equation}
	
	Additionally, the original 6D velocity vector \(\mathbf{u}_t = [\mathbf{v}_t^\top, \boldsymbol{\omega}_t^\top]^\top\) comprises linear velocity \(\mathbf{v}_t\) and angular velocity \(\boldsymbol{\omega}_t\). It is decomposed to construct a new variable suitable for the diffusion process:
	
	\begin{equation}
		\mathbf{d}_t = \left[ \frac{\mathbf{v}_t}{\|\mathbf{v}_t\|}^\top, \frac{\boldsymbol{\omega}_t}{\|\boldsymbol{\omega}_t\|}^\top \right]^\top \in \mathbb{R}^6
	\end{equation}
	
	\begin{equation}
		\mathbf{m}_t = \left[ \|\mathbf{v}_t\|, \|\boldsymbol{\omega}_t\| \right]^\top \in \mathbb{R}^2
	\end{equation}
	where \(\mathbf{d}_t\) is a unit direction vector that encodes the direction of the linear and angular velocity, while \(\mathbf{m}_t\) is a magnitude vector that represents the corresponding speeds.Then we define $\mathbf{action}_t$ as:
	\begin{equation}
		\mathbf{action}_t = [\mathbf{d}_t^\top, \mathbf{m}_t^\top]^\top \in \mathbb{R}^8
	\end{equation}
	
	Using normalized image plane coordinates as model input provides two primary advantages. First, normalized coordinates offer higher geometric resolution compared to pixel coordinates. Second, networks trained on normalized coordinates generalize more effectively across cameras with varying intrinsic parameters.
	
	Because camera velocity is relatively high at the beginning of the servoing process and decreases as the target is approached, directly using \(\mathbf{m}_t\) to compute the loss biases the network toward early-stage data and underweights the critical final phase. To mitigate this issue, \(\mathbf{m}_t\) is mapped to a scalar \(\mathbf{l}\) before being input into the network, as defined by the following mapping:
	
	\begin{equation}
		\mathbf{l}=\mathcal{T}^{-1}(\mathbf{m_{t}})
	\end{equation}
	where $\mathcal{T}(\cdot) = \ln(1 + \exp(\cdot))$, This mapping leverages the logarithmic property to expand velocity magnitudes \(\mathbf{m}_t\) near \(\mathbf{0}\) into the negative infinite range. This transformation significantly stretches the dynamic range of low-speed data, preventing the network from being dominated by large velocities during the early servoing phase and enabling balanced learning across the entire servoing trajectory.
	
	\subsection{DiffusionVS}
	
	In traditional Image-Based Visual Servoing (IBVS), the camera velocity is computed from the difference between the current image feature coordinates \(\mathbf{s_I} = [u, v]^\top\) and the desired feature coordinates \(\mathbf{s^*_I}= [u^*, v^*]^\top\). The error \(\mathbf{e_I}\) is defined as:
	\begin{equation}
		\mathbf{e_I} = \mathbf{s_I} - \mathbf{s^*_I}
	\end{equation}
	
	A classical control law typically employs the pseudo-inverse of the interaction matrix:
	
	\begin{equation}
		\mathbf{v_I} = -\lambda \widehat{\mathbf{L}}_s^{+} \mathbf{e_I}
	\end{equation}
	where \(\mathbf{L}_s\) denotes the interaction matrix (also known as the image Jacobian), and \(\lambda > 0\) is a gain parameter that controls the convergence rate.
	
	The input and output formulation of DiffusionVS closely resembles that of traditional IBVS. Specifically, we define the generation target of the diffusion model as an action sequence $\mathbf{a}_0$ over the next \(H\) time steps:  
	\begin{equation}
		\mathbf{a}_0 = [\mathbf{action}_t, \dots, \mathbf{action}_{t+H-1}] \in \mathbb{R}^{H \times 8}
	\end{equation}
	
	Each action encodes a decoupled representation of the 6D camera velocity. The ground-truth sequence $\mathbf{a}_0$ is derived using a PBVS control law. Specifically, let $\mathbf{P}, \mathbf{P}^* \in \mathbb{R}^{3n}$ denote the stacked current and desired 3D coordinates of $n$ target points, which are acquired directly from the simulation environment to avoid pose estimation errors. 
	
	To map 3D coordinate errors to 6D camera velocities, we define the interaction matrix $\mathbf{L}_{\mathbf{P}_i} \in \mathbb{R}^{3 \times 6}$ for a single point $\mathbf{P}_i = [X_i, Y_i, Z_i]^\top$ as:
	\begin{equation}
		\mathbf{L}_{\mathbf{P}_i} = \begin{bmatrix} -\mathbf{I}_3 & [\mathbf{P}_i]_{\times} \end{bmatrix}
	\end{equation}
	where $\mathbf{I}_3$ is the identity matrix and $[\mathbf{P}_i]_{\times}$ is the skew-symmetric matrix of $\mathbf{P}_i$. By stacking $\mathbf{L}_{\mathbf{P}_i}$ for all $n$ points, we obtain the complete interaction matrix $\mathbf{L}_{\mathbf{P}} \in \mathbb{R}^{3n \times 6}$. The ground-truth camera velocity $\mathbf{a}_0$ is then computed via the proportional control law:
	\begin{equation}
		\mathbf{a}_0 = -\lambda \mathbf{L}_{\mathbf{P}}^+ (\mathbf{P} - \mathbf{P}^*)
	\end{equation}
	where $\lambda > 0$ is the control gain and $\mathbf{L}_{\mathbf{P}}^+$ is the Moore-Penrose pseudo-inverse of $\mathbf{L}_{\mathbf{P}}$.
	During training, the denoising network \(\boldsymbol{\epsilon}_\theta\) takes the noisy action sequence \(\mathbf{a}_k\), the diffusion timestep \(k\), and the observation condition \(\mathbf{cond}\) as input. \(\mathbf{a}_k\) is obtained by progressively adding noise to the clean action sequence according to the DDPM forward process, as shown in the following equation:
	
	\begin{equation}
		\mathbf{a}_k = \sqrt{\bar{\alpha}_k} \, \mathbf{a}_0 + \sqrt{1 - \bar{\alpha}_k} \, \boldsymbol{\epsilon}
	\end{equation}
	where $\boldsymbol{\epsilon} \sim \mathcal{N}(0, \mathbf{I})$. \(\alpha_k\) and \(\bar{\alpha}_k = \prod_{i=1}^k \alpha_i\) are scheduling parameters of the diffusion process.
	
	The model is trained by minimizing the mean squared error between the predicted and true noise in the decoupled action space:
	\begin{equation}
		\mathcal{L}(\theta) = \mathbb{E}_{k, \mathbf{a}_0, \boldsymbol{\epsilon}} \left[ \left\| \boldsymbol{\epsilon} - \boldsymbol{\epsilon}_{\theta}(\mathbf{a}_k, k, \mathbf{cond}) \right\|^2 \right].
	\end{equation}
	
	Once trained, the denoising network can generate appropriate camera velocities via reverse diffusion conditioned on new observations. At inference time, the process starts from pure Gaussian noise \(\mathbf{a}_K \sim \mathcal{N}(\mathbf{0}, \mathbf{I})\). For each denoising step \(k\) (from \(K\) down to \(1\)), the model uses the current observation \(\mathbf{cond}\) to guide the reconstruction of a less noisy action sequence \(\mathbf{a}_{k-1}\) via the following state transition:
	\begin{equation}
		\resizebox{0.4\textwidth}{!}{%
			$\mathbf{a}_{k-1} = \frac{1}{\sqrt{\alpha_k}} \left( \mathbf{a}_k - \frac{1 - \alpha_k}{\sqrt{1 - \bar{\alpha}k}} \boldsymbol{\epsilon}_{\theta}(\mathbf{a}_k, k, \mathbf{cond}) \right) + \sigma_k \mathbf{z}\label{eq:sampling}$
		}
	\end{equation}
	where \(\sigma_k\) is the step-dependent variance (often set to \(\sqrt{(1 - \bar{\alpha}_{k-1})/(1 - \bar{\alpha}_k)} \cdot (1 - \alpha_k)\) for DDPM sampling), and \(\mathbf{z} \sim \mathcal{N}(0, \mathbf{I})\) is standard Gaussian noise (omitted in the final step \(k=1\)). The resulting \(\mathbf{a}_0\) provides a smooth, PBVS-like velocity command sequence without requiring exact depth information.
	
	\subsection{Network Architecture}
	
	\begin{figure*}[!htb]
		\centering
		\includegraphics[width =0.8 \textwidth]{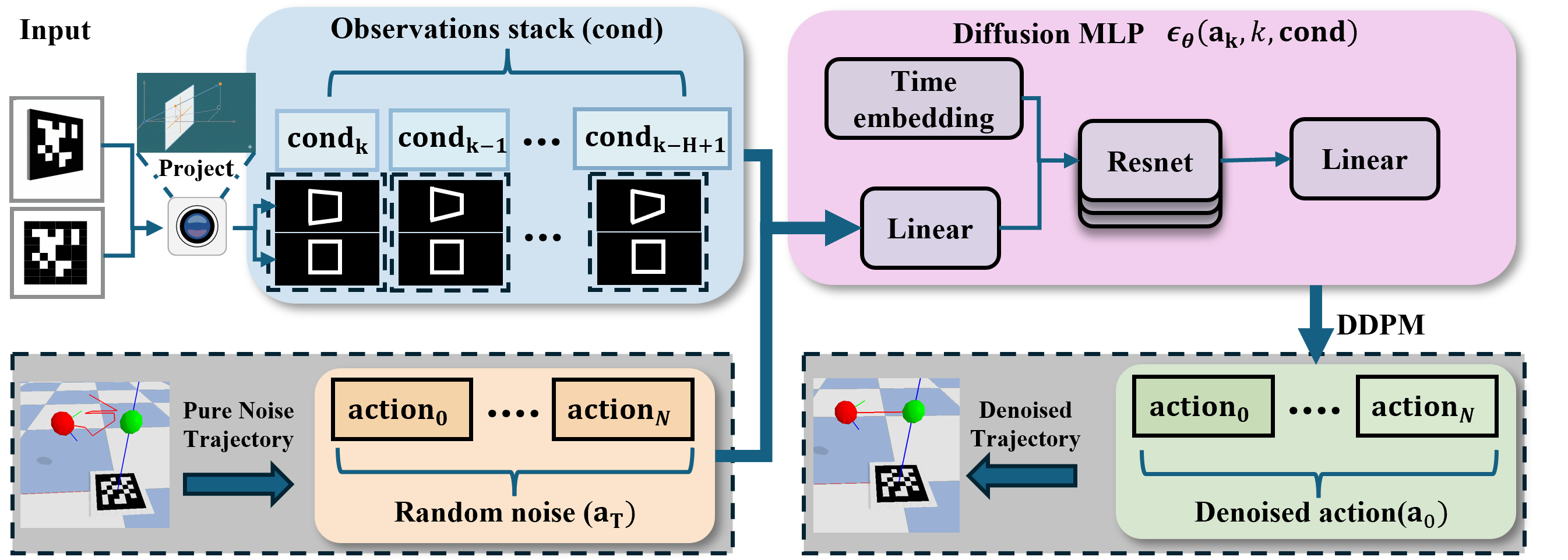}
		\caption{Network architecture of DiffusionVS.}
		\label{network_architecture}
		\vspace{-1em} 
	\end{figure*}
	
	The network architecture employed in this study is intentionally simple, as depicted in Fig.~\ref{network_architecture}. It utilizes a standard multilayer perceptron (MLP)-based conditional diffusion denoising framework. The model receives three inputs: the diffusion timestep \(t\), the noisy action \(\mathbf{a}\), and the conditioning observation \(\mathbf{cond}\).
	
	As illustrated in Fig. 2, the network takes two primary sequences as inputs to incorporate temporal context. The conditioning vector, $\mathbf{cond}$, consists of an observation stack containing normalized servoing points coordinates over a short history of $H$ frames (from step $k-H+1$ to the current step $k$). Concurrently, the network processes an action sequence spanning $N$ timesteps, starting as pure random noise $\mathbf{a}_T$.
	
	These inputs are fed into the Diffusion MLP to predict the noise added during the forward process. First, the noisy action sequence and the observation stack $\mathbf{cond}$ are concatenated and processed through an initial Linear layer (with Mish activation\cite{misra2019mish}) to integrate motion and visual information. Meanwhile, the diffusion timestep $t$ is encoded using a separate Time embedding module.
	
	The integrated input features and the time embeddings are then combined and passed through a stack of MLP-based ResNet blocks. Each block utilizes residual connections and linear layers with Mish nonlinearities to extract deep, hierarchical features. Finally, the processed features are projected back into the action space through a final Linear layer, outputting the predicted noise $\boldsymbol{\epsilon}_\theta$. This prediction directs the iterative denoising process, progressively refining the random noise $\mathbf{a}_T$ into an executable, denoised action trajectory $\mathbf{a}_0$.
	
	\subsection{Online Training with Replay Buffer}
	
	Inspired by recent advances in reinforcement learning \cite{d2022sample}\cite{nikishin2022primacy}, this work adopts an online learning strategy to train the network. In simulation, at each time step, the camera computes a ground-truth expert velocity \(\mathbf{v}_{\text{gt}}\) using the PBVS principle and the true scene state provided by the simulator. Simultaneously, the network generates a student velocity \(\mathbf{v}_{\theta}\), which is used to guide the camera to a new state. The pair \((\mathbf{v}_{\text{gt}}, \mathbf{v}_{\theta})\) along with the corresponding observations is then stored for training.
	
	A primary advantage of online training over static dataset-based approaches is its capacity to expose the network to a broader and more diverse set of data distributions, which significantly enhances robustness. In static datasets, the model observes only trajectories generated by an expert policy. If strong disturbances or noise during real-world servoing cause the system to deviate from this expert distribution, the model lacks the ability to recover. In contrast, online learning allows the network to continuously explore, make mistakes, and self-correct through interaction, thereby gradually learning to manage out-of-distribution states and recover toward the goal.
	\begin{algorithm}[h]
		\caption{DiffusionVS Online Training}
		\label{alg:training_pipeline}
		\DontPrintSemicolon
		\SetNoFillComment
		
		\KwIn{Parallel Envs $\mathcal{E}$, Buffer $\mathcal{D}$, Policy $\pi_\theta$}
		
		\While{not converged}{
			Interact with $N$ envs in parallel: execute $\mathbf{v}_\theta = \pi_\theta(o)$, obtain expert $\mathbf{v}_{gt}$, and store transitions $(o, \mathbf{v}_{gt})$ into $\mathcal{D}$\;
			
			Sample batch $\mathbf{x} \sim \mathcal{D}$, time $t \sim \mathcal{U}(0,1)$, noise $\mathbf{\epsilon} \sim \mathcal{N}(\mathbf{0}, \mathbf{I})$\;
			
			Compute noisy state $\mathbf{z}_t = t \mathbf{x} + (1-t)\mathbf{\epsilon}$ and target $\mathbf{v}_{tar} = (\mathbf{x} - \mathbf{z}_t)/(1-t)$\;
			
			Update $\theta$ to minimize $L = || \pi_\theta(\mathbf{z}_t, t) - \mathbf{v}_{tar} ||^2$\;
		}
	\end{algorithm}
	The complete training pipeline is detailed in Algorithm \ref{alg:training_pipeline}. During training, multiple parallel environments simultaneously generate observation-action tuples $(o, \mathbf{v}_{gt})$, which are stored in a shared replay buffer $\mathcal{D}$. This parallelized collection accelerates training and mitigates online overfitting.
	
	For parameter updates, the network samples mini-batches of expert actions $\mathbf{x}$ from $\mathcal{D}$. Concurrently, it samples a continuous timestep $t \sim \mathcal{U}(0,1)$ (standard uniform distribution) and noise $\boldsymbol{\epsilon} \sim \mathcal{N}(\mathbf{0}, \mathbf{I})$ (standard multivariate Gaussian distribution with zero mean and identity covariance).
	
	To train the diffusion model, the forward process constructs the noisy state $\mathbf{z}_t$ via linear interpolation between the clean data $\mathbf{x}$ and the noise $\boldsymbol{\epsilon}$. Finally, the policy network $\pi_\theta$ is optimized by minimizing the Mean Squared Error (MSE) against the derived learning target $\mathbf{v}_{tar}$.
	
	\section{SIMULATIONS AND EXPERIMENTS}
	
	\subsection{Evaluation Metrics}  
	For the purpose of convenient validation, we utilize an AprilTag calibration board as the visual target; However, the network is not restricted to the specific type of target. We first define the following metrics to evaluate servoing performance:  
	(1) \textbf{TE (Translation Error)}: the Euclidean distance between the final and desired camera positions;  
	(2) \textbf{RE (Rotation Error)}: The angular distance defined as the rotation magnitude in the Axis-Angle representation. It represents the scalar angle required to rotate the camera from its current orientation to the desired orientation around a single fixed axis.
	(3) \textbf{SR (Success Rate)}: the percentage of trials that complete the servoing task. The precise criteria for determining success are specified separately for each experiment.
	
	\subsection{Simulation}  \label{Sim_setup}
	
	A PyBullet-based simulation environment was constructed, comprising a monocular camera and an AprilTag board, as illustrated in Fig.~\ref{fig:simulation_env}. The AprilTag board is fixed in a horizontal orientation and remains stationary throughout each trial, while the camera's initial and target poses are randomly generated.
	
\begin{figure}[htbp]
		\centering
		\includegraphics[width = 0.5\textwidth]{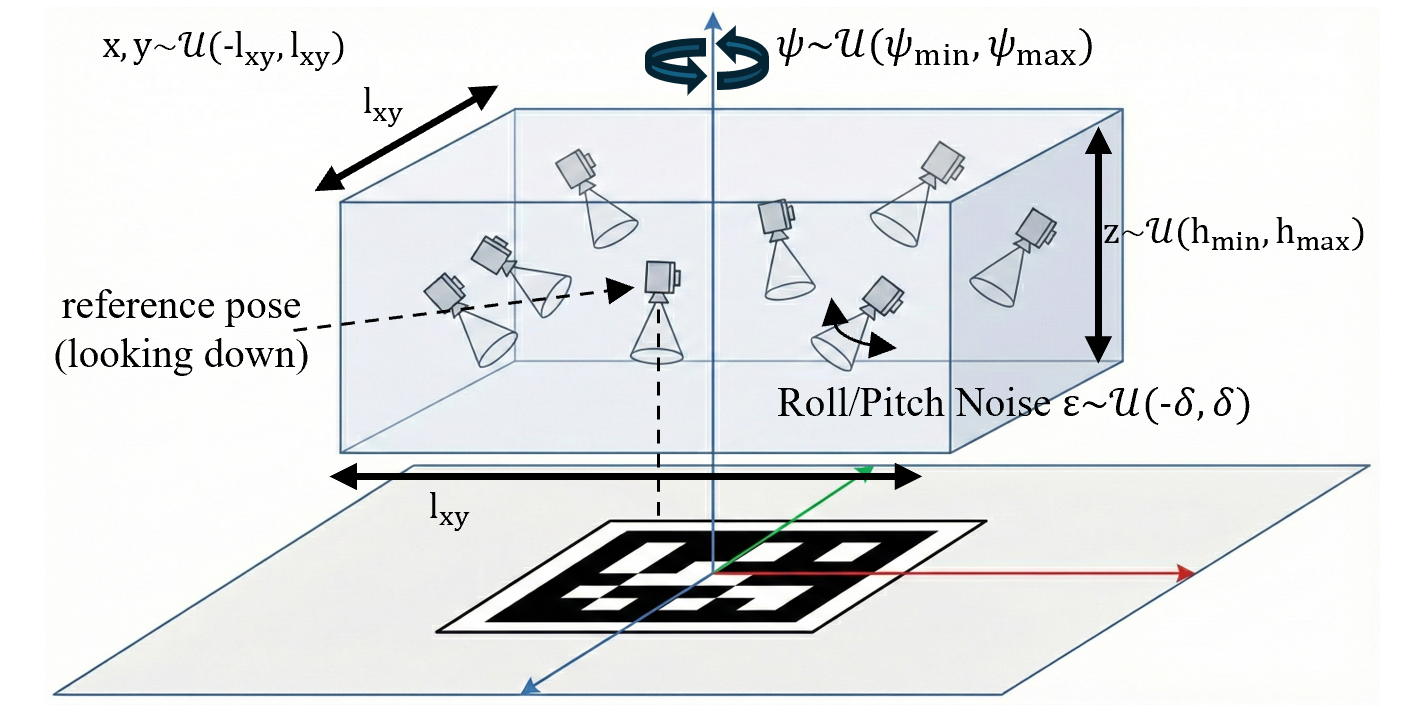}
		\caption{Setup for viewpoint generation in the simulation environment.}
		\label{fig: Sample pose}
		\vspace{-1em} 
	\end{figure}

\begin{figure}[htbp]
		\centering
		\begin{subfigure}{0.23\textwidth} 
			\centering
			\includegraphics[width=0.97\textwidth]{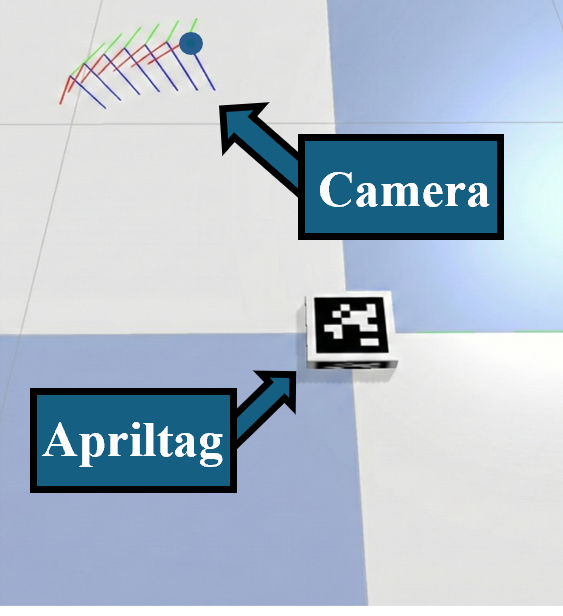} 
			\caption{Simulation environment}
			\label{fig:simulation_env}
		\end{subfigure}
		\hfill 
		\begin{subfigure}{0.23\textwidth} 
			\centering
			\includegraphics[width=\textwidth]{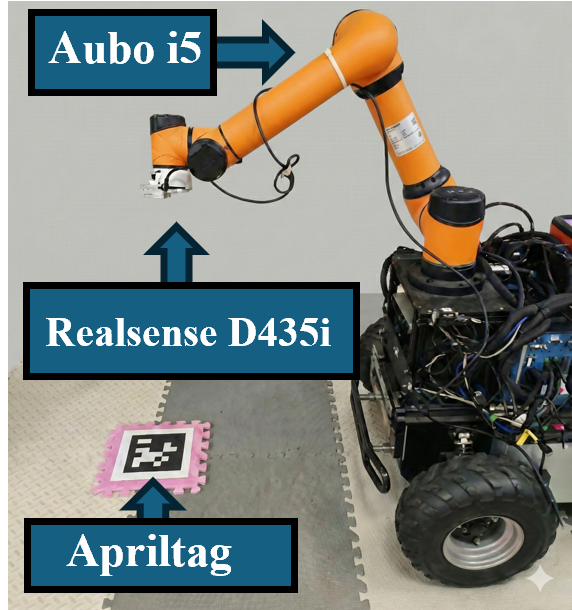}
			\caption{Real robot environment}
			\label{fig:real_robot_env}
		\end{subfigure}
		
		\caption{Evaluation environments}
		\label{fig:main}
		
	\end{figure}

	All sample pose configurations adhere to the random sampling protocol depicted in Fig. \ref{fig: Sample pose}. The position \( p \) is uniformly sampled within a horizontal region centered on the target, with \( x, y \sim \mathcal{U}(-l_{xy}, l_{xy}) \), and within a vertical range \( z \sim \mathcal{U}(h_{\text{min}}, h_{\text{max}}) \). The orientation \( q \) is derived from a top-down reference pose, with uniform noise \( \epsilon \sim \mathcal{U}(-\delta, \delta) \) independently added to the Roll and Pitch axes. The Yaw angle is constrained to \( \psi \sim \mathcal{U}(\psi_{\text{min}}, \psi_{\text{max}}) \). This approach ensures consistent yet diverse viewpoint coverage across all experimental settings.
	
	The sampling of camera poses is constrained in height, preventing the camera from rising excessively above the tag plane. This restriction is essential because, at greater heights, the projected tag corners converge in the normalized image plane, resulting in significant loss of geometric discriminability and hindering the network's ability to learn meaningful spatial relationships. Although the trained model assumes camera heights within a specific range, it can generalize to servoing tasks at different scales, provided that the target points remain sufficiently separated in the image plane. Generalization is achieved by scaling the predicted velocity using an appropriate scale factor.

	The core claim of this work is first validated: in the task of visual servoing, the diffusion-based architecture significantly outperforms conventional regression-based approaches. To clearly illustrate this performance gap, two controlled ablation experiments are conducted using networks with nearly identical parameter counts and architectural components. Specifically, the regression baseline is derived by removing only the timestep embedding module from the DiffusionVS MLP, while all other layers, including the ResNet blocks and feature fusion pathways, remain unchanged. The input for both models consists of normalized AprilTag corner coordinates. However, while the diffusion model predicts noise for the denoising process, the regression model directly outputs camera velocity.
	
	Both models are trained for the same number of iterations under identical conditions and are then evaluated over 100 trials in the same simulated servoing scenarios. A trial is considered successful if the final translation error satisfies \( \text{TE} < 3\,\text{cm} \) and the rotation error satisfies \( \text{RE} < 0.6^\circ \). The quantitative results are summarized in Table~\ref{tab:diffusion_vs_regression_styled}.
	
	\begin{table}[htbp]
		\centering
		\caption{Quantitative comparison between Regression and Diffusion approaches.}
		\label{tab:diffusion_vs_regression_styled}
		
		\newcolumntype{Y}{>{\centering\arraybackslash}X}
		
		\begin{tabularx}{\columnwidth}{l l Y Y Y}
			\toprule
			\multirow{2}{*}{\textbf{Category}} & \multirow{2}{*}{\textbf{Method}} & \textbf{SR} & \textbf{TE} & \textbf{RE} \\
			& & ($\%$, $\uparrow$) & (cm, $\downarrow$) & ($^{\circ}$, $\downarrow$) \\
			\midrule
			\multirow{2}{*}{\textbf{Regression}} & wo/ Pose & 31 & 1.81 & 1.83 \\
			& w/ Pose & 67 & 1.51 & 1.50 \\
			\midrule
			\multirow{2}{*}{\textbf{Diffusion}}
			& \textbf{wo/ Pose} & \textbf{100} & \textbf{0.54} & \textbf{0.53} \\ 
			& w/ Pose & 97 & 0.79 & 0.86 \\
			\bottomrule
			\multicolumn{5}{p{\dimexpr\columnwidth-2\tabcolsep\relax}}{\footnotesize \textit{* Averaged 100 trials for each group. Success: TE $< 5$cm, RE $< 3^{\circ}$.}} \\
		\end{tabularx}
		\vspace{-1em} 
	\end{table}
	
	All experimental groups were trained in identical simulation environments for an equivalent duration. As indicated in Table~\ref{tab:diffusion_vs_regression_styled}, the diffusion-based architecture consistently achieves strong performance, irrespective of the availability of ground-truth camera pose as input. In contrast, the regression-based model performs significantly worse than the diffusion-based counterpart, even when pose information is provided. When pose input is omitted, its success rate falls below 50\%. This pronounced performance disparity highlights the superiority of the diffusion framework for visual servoing tasks.
	
	For existing state-of-the-art (SOTA) end-to-end regression-based visual servoing networks, integrating a diffusion framework can further improve servoing performance. To demonstrate this enhancement, the GraphVS network from the CNS framework\cite{Chen2023CNSCE} was modified by replacing its original decoder with the Diffusion MLP. The performance of this enhanced variant, referred to as GraphDiffVS, was then compared to the original GraphVS under the same three experimental settings detailed in Table \ref{tab:pose_params}. Evaluating across these multiple scenarios is essential to fully expose the performance advantages of the diffusion mechanism against the strong GraphVS baseline. The results are presented in Table \ref{tab:performance_comparison}:
	
	\begin{table}[h]
		\centering
		\caption{Parameters for Random Pose Generation across Three Experimental Groups}
		\label{tab:pose_params}
		\begin{tabular}{l c c c c}
			\toprule
			\textbf{Group} & $l_{xy}$ & [$h_{min}, h_{max}$] & $\delta$ & $[\psi_{min},\psi_{max}]$ \\
			\midrule
			\textbf{Set A} & $0.3$\,m & $[0.4, 0.8]$\,m & $0.2$\,rad & $[-\pi, \pi]$ \\
			\textbf{Set B } & $0.5$\,m & $[0.4, 1.5]$\,m & $0.5$\,rad & $[-\pi, \pi]$ \\
			\textbf{Set C } & $1.0$\,m & $[0.4, 2.0]$\,m & $1.0$\,rad & $[-\pi, \pi]$ \\
			\bottomrule
		\end{tabular}
	\end{table}

	\begin{table}[htbp]
		\centering
		\caption{Quantitative comparison between GraphVS and GraspDiffVS.}
		\label{tab:performance_comparison}
		
		\newcolumntype{Y}{>{\centering\arraybackslash}X}
		
		\begin{tabularx}{\columnwidth}{l l Y Y Y}
			\toprule
			\multirow{2}{*}{\textbf{Setting}} & \multirow{2}{*}{\textbf{Method}} & \textbf{SR} & \textbf{TE} & \textbf{RE} \\
			& & ($\%$, $\uparrow$) & (cm, $\downarrow$) & ($^{\circ}$, $\downarrow$) \\
			\midrule
			\multirow{2}{*}{\textbf{Set A}} & GraphVS & 98 & 0.22 & 0.21 \\
			& \textbf{GraphDiffVS} & \textbf{100} & \textbf{0.19} & \textbf{0.14} \\
			\midrule
			\multirow{2}{*}{\textbf{Set B}} & GraphVS & 65 & 0.44 & 0.26 \\
			& \textbf{GraphDiffVS} & \textbf{96} & \textbf{0.27} & \textbf{0.18} \\
			\midrule
			\multirow{2}{*}{\textbf{Set C}} & GraphVS & 21 & 0.64 & \textbf{0.33} \\
			& \textbf{GraphDiffVS} & \textbf{52} & \textbf{0.63} & 0.35 \\
			\bottomrule
			\multicolumn{5}{p{\dimexpr\columnwidth-2\tabcolsep\relax}}{\footnotesize \textit{* Averaged 100 trials for each group. Success: TE $< 1$cm, RE $< 1^{\circ}$.}} \\
		\end{tabularx}
		\vspace{-1em} 
	\end{table}
	
	The results indicate that augmenting GraphVS with a diffusion-based decoder consistently improves performance across all three task difficulty levels. These findings demonstrate that the performance gains associated with the diffusion framework extend beyond the previously introduced Diffusion MLP architecture and can be effectively applied to more complex regression-based visual servoing networks.
	
	Table \ref{tab:computational_time} summarizes the computational efficiency of the proposed framework. The perception components are highly efficient (Encoder: 1.24 ms, Backbone: 1.39 ms), and the Diffusion Policy takes just 7.58 ms. With a total execution time of ~10.2 ms per step (nearly 100 Hz).
		
	\begin{table}[htbp]
		\centering
		\caption{Computational Efficiency of the GraphDiffVS.}
		\label{tab:computational_time}
		
		\newcolumntype{Y}{>{\centering\arraybackslash}X}
		
		\begin{tabularx}{\columnwidth}{l Y Y Y}
			\toprule
			& \textbf{Encoder} & \textbf{Backbone} & \textbf{Diffusion} \\
			\midrule
			\textbf{TpIT} (ms) & 1.24 & 1.39 & 7.58 \\
			\bottomrule
			\multicolumn{4}{p{\dimexpr\columnwidth-2\tabcolsep\relax}}{\footnotesize \textit{* TpIT: Time per Iteration. The Diffusion module includes both direction and magnitude inference.}} \\
		\end{tabularx}
		\vspace{-1em} 
	\end{table}
	
	\subsection{Real-World Experiment} \label{Real setup} 
	
	The proposed method is validated on a fixed-base AUBO-i5 6-DOF robotic arm (±0.02 mm repeatability) equipped with an eye-in-hand calibrated RGB camera (Fig.~\ref{fig:real_robot_env}). Initial and target camera poses are randomly sampled and perturbed within a hemispherical workspace (radius 0.6 m to 0.8 m) centered above a ground-mounted AprilTag. During operation, the network takes normalized AprilTag corner coordinates—extracted from raw RGB images—as input and outputs camera-frame velocity commands to actuate the robot. A trial is considered successful if the total pixel-wise error of the corners is under 15 pixels after 100 servoing steps. Table~\ref{tab:real_robot_contrast} compares the real-world performance of the diffusion-based model against the regression baseline under these conditions.
	
	\begin{table}[htbp]
		\centering
		\caption{Quantitative comparison of performance on the real robot experiment.}
		\label{tab:real_robot_contrast}
		
		\newcolumntype{Y}{>{\centering\arraybackslash}X}
		
		\begin{tabularx}{\columnwidth}{l l Y Y Y}
			\toprule
			& \textbf{Method} & \textbf{SR} ($\%$, $\uparrow$) & \textbf{TE} (cm, $\downarrow$) & \textbf{RE} ($^{\circ}$, $\downarrow$) \\
			\midrule
			& Regression & 0.0 & -- & -- \\
			& \textbf{Diffusion (Ours)} & \textbf{93.3} & \textbf{6.17} & \textbf{3.76} \\
			\bottomrule
			\multicolumn{5}{p{\dimexpr\columnwidth-2\tabcolsep\relax}}{\footnotesize \textit{* Averaged 30 trials for each group. Success: Pixel error $<$ 15 pixels.}} \\
		\end{tabularx}
	\end{table}
	
	Table~\ref{tab:real_robot_contrast} indicates that the diffusion-based model exhibits strong sim-to-real transferability, achieving reliable and accurate servoing on the physical robot. In contrast, the regression-based model fails when deployed in the real world, exhibiting unstable or divergent behavior. This failure is primarily due to the higher camera observation noise present in physical experiments. On one hand, the regression-based model is highly sensitive to this noise, resulting in indecisive control actions that cause the robot to oscillate around the target without converging. On the other hand, once the robot deviates from the correct trajectory, the model sometimes suffers from severe cascading errors, which further exacerbate the divergence. These vulnerabilities underscore its poor generalization beyond simulated conditions.
	
	\subsection{Ablation Experiment}
	
	To assess whether online training substantially enriches the diversity of training data, a comparative experiment was conducted between the proposed continuous Online Training approach and an offline training baseline using a static dataset (referred to as Static in Table~\ref{tab:online_vs_static_simple} and Table~\ref{tab:chart_online_static_Contrast_real}). For a fair comparison, both approaches utilized the same total amount of data: the online process collected 2,400,000 interaction frames, matching the 2,400,000 pre-recorded frames in the static dataset. The network architecture and data preprocessing pipeline remained identical across both settings. Furthermore, to ensure the static baseline was sufficiently trained and to investigate the impact of training duration, three distinct static groups were evaluated at varying numbers of epochs (abbreviated as ep.): 500, 1000, and 2000 epochs.
		
	Notably, restricting both methods to the same data volume inherently disadvantages online training. Under this constraint, the online model receives significantly fewer gradient updates per sample, effectively limiting its primary advantage of exploiting an unlimited, continuous data stream. The resulting models are evaluated under identical conditions in both simulation (Set A configuration in Table \ref{tab:pose_params}) and real-world experiments. The simulation results are summarized in Table~\ref{tab:online_vs_static_simple}, and the real-robot results are presented in Table \ref{tab:chart_online_static_Contrast_real}. The experimental setups for both studies are identical to those described in Sections \ref{Sim_setup} and \ref{Real setup}.
	
	\begin{table}[htbp]
		\centering
		\caption{Performance contrast between Online Training and Static Training baselines.}
		\label{tab:online_vs_static_simple}
		
		\newcolumntype{Y}{>{\centering\arraybackslash}X}
		
		\begin{tabularx}{\columnwidth}{l Y Y Y}
			\toprule
			\textbf{Method} & \textbf{SR} ($\%$, $\uparrow$) & \textbf{TE} (cm, $\downarrow$) & \textbf{RE} ($^{\circ}$, $\downarrow$) \\
			\midrule
			\textbf{Online Training} & \textbf{77} & \textbf{0.37} & 0.54 \\
			\midrule
			Static (500 ep.)  & 30  & 2.29 & 1.70 \\
			Static (1000 ep.) & 53  & 3.01 & 0.89 \\
			Static (2000 ep.) & 4 & 4.62 & \textbf{0.39} \\
			\bottomrule
			\multicolumn{4}{p{\dimexpr\columnwidth-2\tabcolsep\relax}}{\footnotesize \textit{* Averaged 100 trials for each group. Success: TE $< 5$cm, RE $< 3^{\circ}$.}} \\
		\end{tabularx}
		\vspace{-1em}
	\end{table}

	The results indicate that the success rate of online training does not exceed 90\%. This outcome is attributed to the model in this ablation study being trained on only 2.4 million frames, in contrast to the models in previous experiments that benefited from extensive interaction. Consequently, the model was not fully converged.
	
	Despite this disadvantageous setup, online training received significantly fewer effective gradient updates compared to offline training. The results further demonstrate that the simulation group begins to exhibit performance degradation due to overfitting after 1000 epochs. Furthermore, even at its peak performance around the 1000-epoch mark, the success rate remains below 30\%, which indicates fundamental limitations of the offline training paradigm for this task.
	
	\begin{table}[htbp]
		\centering
		\caption{Contrast between online training and static training (Real Robot).}
		\label{tab:chart_online_static_Contrast_real}
		
		\newcolumntype{Y}{>{\centering\arraybackslash}X}
		
		\begin{tabularx}{\columnwidth}{l Y Y Y}
			\toprule
			\textbf{Method} & \textbf{SR} ($\%$, $\uparrow$) & \textbf{TE} (cm, $\downarrow$) & \textbf{RE} ($^{\circ}$, $\downarrow$) \\
			\midrule
			\textbf{Online Training} & \textbf{76.7} & \textbf{16.82} & \textbf{10.78} \\
			\midrule
			Static (500 ep.)  & 0 & -- & -- \\
			Static (1000 ep.) & 0 & -- & -- \\
			Static (2000 ep.) & 0 & -- & -- \\
			\bottomrule
			\multicolumn{4}{p{\dimexpr\columnwidth-2\tabcolsep\relax}}{\footnotesize \textit{* Averaged 30 trials for each group. Success: pixel error $<$ 15 pixels.}} \\
		\end{tabularx}
	\end{table}

	As shown in Table~\ref{tab:chart_online_static_Contrast_real}, the online-trained model in real-world experiments demonstrates a slight decrease in success rate and increased error compared to previous physical trials, primarily attributable to limited training (2.4 million frames). Nevertheless, it significantly outperforms the offline-trained model.
	
	The offline model fails in real-world tests primarily in two ways. First, it may begin with the correct servoing direction but then diverge at an intermediate stage and fail to recover, likely due to the static dataset lacking examples of corrective actions following trajectory drift. This limitation is not apparent in the simulation because of the environment's simplified and deterministic nature. However, the more variable observation distribution in real-world settings reveals the model’s inability to manage out-of-distribution states. Second, the offline model sometimes approaches the target without major deviation but displays large, persistent oscillations near the goal pose, mirroring its behavior in simulation.
	
	In contrast, the online-trained model consistently achieves stable and smooth convergence in physical experiments, demonstrating superior sim-to-real transfer capability and robustness to real-world visual and dynamic perturbations.
	
	In summary, online training demonstrates superior effectiveness for visual servoing tasks. Although both approaches utilize the same amount of data, the resulting models exhibit markedly different performance. This highlights a key insight: expert datasets collected offline contain substantial information redundancy and lack the diversity required for robust policy learning. Conversely, online training continuously generates rich, task-relevant, and diverse experiences through interaction, enabling the model to acquire more adaptable and reliable control strategies. For dynamic control tasks such as visual servoing, data diversity and coverage of corrective behaviors are more critical than data volume alone.
	
	\section{CONCLUSIONS}
	
	This work introduces DiffusionVS, a novel visual servoing method based on Diffusion Policy. DiffusionVS produces high-quality servoing trajectories using only normalized pixel coordinates as input, effectively combining the robustness to depth uncertainty of traditional IBVS with the smooth, globally consistent trajectories of PBVS. To address the sim-to-real gap, an online training framework is developed that continuously enriches the training data through interactive experience collection. This approach significantly increases data diversity and substantially improves real-world performance.
	
	Extensive simulations and experiments, conducted both in simulation and on a physical AUBO-i5 robotic arm, demonstrate that DiffusionVS consistently outperforms conventional regression-based end-to-end models, which often struggle with generalization and trajectory stability. Furthermore, results indicate that online training yields substantially higher success rates and robustness compared to learning from static datasets.
	
	The proposed approach is highly modular and readily transferable to existing engineering systems. By retaining a standard backbone network and integrating diffusion-specific components, such as the time-step embedding and the iterative denoising training pipeline, regression-based architectures can be upgraded to leverage the expressive power and robustness of diffusion-based control.
	
	\bibliographystyle{IEEEtran}
	
	\bibliography{refs}          

@article{chaumette2006visual,
	title={Visual servo control. I. Basic approaches},
	author={Chaumette, Fran{\c{c}}ois and Hutchinson, Seth},
	journal={IEEE Robotics \& Automation Magazine},
	volume={13},
	number={4},
	pages={82--90},
	year={2006},
	publisher={IEEE}
}

@article{hutchinson1996tutorial,
	title={A tutorial on visual servo control},
	author={Hutchinson, Seth and Hager, Gregory D and Corke, Peter I},
	journal={IEEE transactions on robotics and automation},
	volume={12},
	number={5},
	pages={651--670},
	year={1996},
	publisher={IEEE}
}

@article{wilson1996relative,
	title={Relative end-effector control using cartesian position based visual servoing},
	author={Wilson, William J and Williams Hulls, Carol C and Bell, Graham S},
	journal={IEEE Transactions on Robotics and Automation},
	volume={12},
	number={5},
	pages={684--696},
	year={1996},
	publisher={IEEE}
}

@article{espiau1992new,
	title={A new approach to visual servoing in robotics},
	author={Espiau, Bernard and Chaumette, Fran{\c{c}}ois and Rives, Patrick},
	journal={IEEE Transactions on Robotics and Automation},
	volume={8},
	number={3},
	pages={313--326},
	year={1992},
	publisher={IEEE}
}

@inproceedings{chaumette1998potential,
	title={Potential problems of stability and convergence in image-based and position-based visual servoing},
	author={Chaumette, Fran{\c{c}}ois},
	booktitle={The Confluence of Vision and Control},
	pages={66--78},
	year={1998},
	organization={Springer}
}

@inproceedings{saxena2017exploring,
	title={Exploring convolutional networks for end-to-end visual servoing},
	author={Saxena, A. and Pandya, H. and Kumar, G. and Gaud, A. and Krishna, K. M.},
	booktitle={2017 IEEE International Conference on Robotics and Automation (ICRA)},
	pages={3817--3823},
	year={2017},
	organization={IEEE}
}

@inproceedings{bateux2018training,
	title={Training deep neural networks for visual servoing},
	author={Bateux, Quentin and Marchand, Eric and Leitner, J{\"u}rgen and Chaumette, Fran{\c{c}}ois and Corke, Peter},
	booktitle={2018 IEEE International Conference on Robotics and Automation (ICRA)},
	pages={3307--3314},
	year={2018},
	organization={IEEE}
}

@inproceedings{yu2019siamese,
	title={Siamese convolutional neural network for sub-millimeter-accurate camera pose estimation and visual servoing},
	author={Yu, C. and Cai, Z. and Pham, H. and Pham, Q.-C.},
	booktitle={2019 IEEE/RSJ International Conference on Intelligent Robots and Systems (IROS)},
	pages={935--941},
	year={2019},
	organization={IEEE}
}

@inproceedings{felton2021siamese,
	title={Siame-SE (3): Regression in SE (3) for end-to-end visual servoing},
	author={Felton, S. and Fromont, E. and Marchand, E.},
	booktitle={2021 IEEE International Conference on Robotics and Automation (ICRA)},
	pages={14454--14460},
	year={2021},
	organization={IEEE}
}

@inproceedings{detone2018superpoint,
	title={Superpoint: Self-supervised interest point detection and description},
	author={DeTone, Daniel and Malisiewicz, Tomasz and Rabinovich, Andrew},
	booktitle={Proceedings of the IEEE conference on computer vision and pattern recognition workshops},
	pages={224--236},
	year={2018}
}

@inproceedings{sarlin2020superglue,
	title={Superglue: Learning feature matching with graph neural networks},
	author={Sarlin, Paul-Edouard and DeTone, Daniel and Malisiewicz, Tomasz and Rabinovich, Andrew},
	booktitle={Proceedings of the IEEE/CVF conference on computer vision and pattern recognition},
	pages={4938--4947},
	year={2020}
}

@inproceedings{ni2023pats,
	title={Pats: Patch area transportation with subdivision for local feature matching},
	author={Ni, Junjie and Li, Yijin and Huang, Zhaoyang and Li, Hongsheng and Bao, Hujun and Cui, Zhaopeng and Zhang, Guofeng},
	booktitle={Proceedings of the IEEE/CVF conference on computer vision and pattern recognition},
	pages={17776--17786},
	year={2023}
}

@inproceedings{felton2021siame,
	title={Siame-se (3): regression in se (3) for end-to-end visual servoing},
	author={Felton, Samuel and Fromont, Elisa and Marchand, Eric},
	booktitle={2021 IEEE International Conference on Robotics and Automation (ICRA)},
	pages={14454--14460},
	year={2021},
	organization={IEEE}
}

@article{Chen2023CNSCE,
	title={CNS: Correspondence Encoded Neural Image Servo Policy},
	author={An-Jen Chen and Hongxiang Yu and Yue Wang and Rong Xiong},
	journal={2024 IEEE International Conference on Robotics and Automation (ICRA)},
	year={2023},
	pages={17410-17416},
	url={https://api.semanticscholar.org/CorpusID:262044993}
}

@article{Chen2025CNSv2PC,
	title={CNSv2: Probabilistic Correspondence Encoded Neural Image Servo},
	author={An-Jen Chen and Hongxiang Yu and Shuxin Li and Yuxi Chen and Zhongxiang Zhou and Wentao Sun and Rong Xiong and Yue Wang},
	journal={2025 IEEE International Conference on Robotics and Automation (ICRA)},
	year={2025},
	pages={15516-15522},
	url={https://api.semanticscholar.org/CorpusID:276742454}
}

@article{allibert2010predictive,
	title={Predictive control for constrained image-based visual servoing},
	author={Allibert, Guillaume and Courtial, Estelle and Chaumette, Fran{\c{c}}ois},
	journal={IEEE Transactions on Robotics},
	volume={26},
	number={5},
	pages={933--939},
	year={2010},
	publisher={IEEE}
}

@inproceedings{kermorgant2011combining,
	title={Combining IBVS and PBVS to ensure the visibility constraint},
	author={Kermorgant, Olivier and Chaumette, Fran{\c{c}}ois},
	booktitle={2011 IEEE/RSJ International Conference on Intelligent Robots and Systems},
	pages={2849--2854},
	year={2011},
	organization={IEEE}
}

@article{jin2021policy,
	title={Policy-based deep reinforcement learning for visual servoing control of mobile robots with visibility constraints},
	author={Jin, Zhehao and Wu, Jinhui and Liu, Andong and Zhang, Wen-An and Yu, Li},
	journal={IEEE Transactions on Industrial Electronics},
	volume={69},
	number={2},
	pages={1898--1908},
	year={2021},
	publisher={IEEE}
}

@incollection{chaumette2007potential,
	title={Potential problems of stability and convergence in image-based and position-based visual servoing},
	author={Chaumette, Francois},
	booktitle={The confluence of vision and control},
	pages={66--78},
	year={2007},
	publisher={Springer}
}

@article{gans2007stable,
	title={Stable visual servoing through hybrid switched-system control},
	author={Gans, Nicholas R and Hutchinson, Seth A},
	journal={IEEE Transactions on Robotics},
	volume={23},
	number={3},
	pages={530--540},
	year={2007},
	publisher={IEEE}
}

@article{malis20022,
	title={2 1/2 D visual servoing},
	author={Malis, Ezio and Chaumette, Francois and Boudet, Sylvie},
	journal={IEEE Transactions on Robotics and Automation},
	volume={15},
	number={2},
	pages={238--250},
	year={2002},
	publisher={IEEE}
}

@article{chi2025diffusion,
	title={Diffusion policy: Visuomotor policy learning via action diffusion},
	author={Chi, Cheng and Xu, Zhenjia and Feng, Siyuan and Cousineau, Eric and Du, Yilun and Burchfiel, Benjamin and Tedrake, Russ and Song, Shuran},
	journal={The International Journal of Robotics Research},
	volume={44},
	number={10-11},
	pages={1684--1704},
	year={2025},
	publisher={Sage Publications Sage UK: London, England}
}

@article{dhariwal2021diffusion,
	title={Diffusion models beat gans on image synthesis},
	author={Dhariwal, Prafulla and Nichol, Alexander},
	journal={Advances in neural information processing systems},
	volume={34},
	pages={8780--8794},
	year={2021}
}

@inproceedings{maze2023diffusion,
	title={Diffusion models beat gans on topology optimization},
	author={Maz{\'e}, Fran{\c{c}}ois and Ahmed, Faez},
	booktitle={Proceedings of the AAAI conference on artificial intelligence},
	volume={37},
	number={8},
	pages={9108--9116},
	year={2023}
}

@ARTICLE{10158789,
	author={Yu, Hongxiang and Chen, Anzhe and Xu, Kechun and Zhou, Zhongxiang and Jing, Wei and Wang, Yue and Xiong, Rong},
	journal={IEEE Robotics and Automation Letters}, 
	title={A Hyper-Network Based End-to-End Visual Servoing With Arbitrary Desired Poses}, 
	year={2023},
	volume={8},
	number={8},
	pages={4769-4776},
	doi={10.1109/LRA.2023.3288382}}

@article{felton2022visual,
	title={Visual servoing in autoencoder latent space},
	author={Felton, Samuel and Brault, Pascal and Fromont, Elisa and Marchand, Eric},
	journal={IEEE Robotics and Automation Letters},
	volume={7},
	number={2},
	pages={3234--3241},
	year={2022},
	publisher={IEEE}
}

@article{urbanikova2023arguing,
	title={Arguing about the essence of public service in public service media: a case study of a newsroom conflict at Slovak RTVS},
	author={Urb{\'a}nikov{\'a}, Mar{\'\i}na},
	journal={Journalism Studies},
	volume={24},
	number={10},
	pages={1352--1374},
	year={2023},
	publisher={Taylor \& Francis}
}

@inproceedings{d2022sample,
	title={Sample-efficient reinforcement learning by breaking the replay ratio barrier},
	author={D'Oro, Pierluca and Schwarzer, Max and Nikishin, Evgenii and Bacon, Pierre-Luc and Bellemare, Marc G and Courville, Aaron},
	booktitle={Deep Reinforcement Learning Workshop NeurIPS 2022},
	year={2022}
}

@inproceedings{nikishin2022primacy,
	title={The primacy bias in deep reinforcement learning},
	author={Nikishin, Evgenii and Schwarzer, Max and D’Oro, Pierluca and Bacon, Pierre-Luc and Courville, Aaron},
	booktitle={International conference on machine learning},
	pages={16828--16847},
	year={2022},
	organization={PMLR}
}

@article{he2026cibvs,
	author  = {He, Rui and Xuan, Yuan and Cui, Hongkang and Li, Peng and Chen, Haoyao},
	title   = {CIBVS: Continuous Image-Based Visual Servoing against Visual Signal Loss},
	journal = {IEEE/ASME Transactions on Mechatronics},
	year    = {2026},
	note    = {accepted},
	doi     = {10.1109/TMECH.2026.3657797} 
}

@article{misra2019mish,
	title={Mish: A self regularized non-monotonic activation function},
	author={Misra, Diganta},
	journal={arXiv preprint arXiv:1908.08681},
	year={2019}
}
	
\end{document}